\def\year{2020}\relax
\documentclass[letterpaper]{article} 
\usepackage{aaai20}  
\usepackage{times}  
\usepackage{helvet} 
\usepackage{courier}  
\usepackage[hyphens]{url}  
\usepackage{graphicx} 
\usepackage{graphicx}  
\newcommand{\thistitle}{Presentation and Analysis of a Multimodal Dataset for Grounded Language Learning}

\usepackage{booktabs}
\usepackage{comment}
\usepackage{multirow}
\usepackage{mathtools}
 
\usepackage[utf8]{inputenc}
\usepackage{pgfplots}
\usepackage{amsfonts}
\usepackage{amssymb}

\DeclareUnicodeCharacter{2212}{−}
\usepgfplotslibrary{groupplots,dateplot}
\usetikzlibrary{patterns,shapes.arrows}
\pgfplotsset{compat=newest}

\usepackage{todonotes} 

\usepackage{graphicx}
\usepackage{url}
\usepackage{subcaption}
\usepackage[inline]{enumitem}
\usepackage{cleveref}

\newcommand{\gld}{GoLD}

\newcommand{\citet}[1]{\citeauthor{#1} \shortcite{#1}}
\newcommand{\citep}{\cite}

\usepackage[normalem]{ulem} 

\title{\thistitle}

\author{Patrick Jenkins, Rishabh Sachdeva, Gaoussou Youssouf Kebe, Padraig Higgins, Kasra \\ \Large \textbf{Darvish, Edward Raff,\textsuperscript{\rm 1} Don Engel, John Winder,\textsuperscript{\rm 2} Francis Ferraro, Cynthia Matuszek}\\ 
 University of Maryland, Baltimore County,
 \textsuperscript{\rm 1}Booz Allen Hamilton, 
 \textsuperscript{\rm 2}Johns Hopkins Applied Physics Laboratory\\
 1000 Hilltop Circle, Baltimore, MD, 21250\\
 pjenk1,rishabs1,mb88814,phiggin1,kasradarvish,edraff1,donengel,jwinder1,ferraro,cmat@umbc.edu 
 }

\begin{document}

\maketitle

\begin{abstract}
Grounded language acquisition --- learning how language-based interactions refer to the world around them --- is a major area of research in robotics, NLP, and HCI. In practice the data used for learning consists almost entirely of textual descriptions, which tend to be cleaner, clearer, and more grammatical than actual human interactions. In this work, we present the Grounded Language Dataset (\gld), a multimodal dataset of common household objects described by people using either spoken or written language. We analyze the differences and present an experiment showing how the different modalities affect language learning from human input. This will enable researchers studying the intersection of robotics, NLP, and HCI to better investigate how the multiple modalities of image, text, and speech interact, as well as how differences in the vernacular of these modalities impact results. 
\end{abstract}

\section{Introduction}


Grounded language acquisition is the process of learning language as it relates to the world---how concepts in language refer to objects, tasks, and environments~\cite{mooney2008learning}. \textit{Embodied} language learning specifically is a significant field of research in NLP, machine learning, and robotics. There are many ways in which robots learn grounded language~\cite{chen2008learning,branavan2009reinforcement,tellex2011understanding,thomason2015learning,thomason2018guiding,Matuszek2018LearningGL,Anderson2018VisionandLanguageNI,das2018embodied,chevalierboisvert2019babyai,hu2019you,vanzo2020grounded}, but they all require either multimodal data or natural language data---usually both.

A significant goal of modern robotics research is the development of robots that can operate in human-centric environments. Examples include domestic service robots (DSRs) that handle common household tasks such as cooking, cleaning, and caretaking~\cite{beckerle2017human}, robots for elder care~\cite{bedaf2014overview}, assistive robotics for providing support to people with disabilities~\cite{Chen2013RobotsFH}, and rehabilitation robotics~\cite{kubota2020jessie}. To be useful for non-specialists, such robots will require easy-to-use interfaces~\cite{beer2012domesticated}. Spoken natural language is an appropriate interface for such systems: it is natural, expressive, and widely understood, as shown by the proliferation of natural language-based home devices\cite{haeb2019speech}. To have a robotic system flexibly understand language in dynamic settings and realize it in physical, goal-oriented behaviors, it is necessary to ground linguistic and perceptual inputs to a learned representation of knowledge tied to actions.

Current approaches to grounded language learning require data in both the perceptual (``grounded'') and linguistic domains. While existing datasets have been used for this purpose~\cite{Johnson2016clevr,krishna2017visual,das2018embodied,nguyen2019help,thomason2019vision}, the language component is almost always derived from either textual input or manually transcribed speech~\cite{matuszek2014learning,tellex2011understanding}. In practice, robots are likely to need to operate on imperfectly understood spoken language. To that end, we present the \textbf{G}r\textbf{o}unded \textbf{L}anguage \textbf{D}ataset (\gld), which contains images of common household objects and their description in multiple formats: text, speech (audio), and automatically recognized speech derived from the audio files. We present experiments that demonstrate the utility of this dataset for grounded language learning.

\begin{figure}[]
\centering
\vspace{2ex}\resizebox{\columnwidth}{!}{%
\begin{tabular}{l|c|c|c}
 & \textbf{\begin{tabular}[c]{@{}c@{}}Typed\\ description\end{tabular}} & \textbf{\begin{tabular}[c]{@{}c@{}}Spoken\\ description\end{tabular}} & \textbf{\begin{tabular}[c]{@{}c@{}}Google\\ ASR\end{tabular}} \\ \hline
\multirow{2}{*}{\includegraphics[width=0.13\columnwidth]{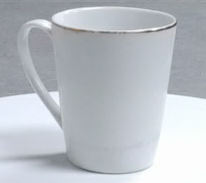}} & \multicolumn{1}{l|}{\includegraphics[width=0.2\columnwidth]{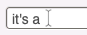}} & \multicolumn{1}{l|}{\includegraphics[width=0.2\columnwidth]{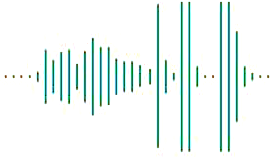}} & \multicolumn{1}{l}{\includegraphics[width=0.2\columnwidth]{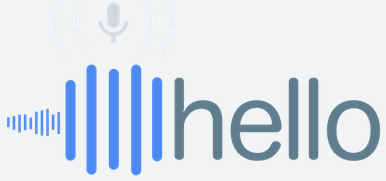}} \\ \cline{2-4} 
 & \begin{tabular}[c]{@{}c@{}}it's a\\ coffee mug\end{tabular} & \begin{tabular}[c]{@{}c@{}}``There is a white\\ coffee mug.''\end{tabular} & \begin{tabular}[c]{@{}c@{}}Arizona white\\ coffee mug\end{tabular}
\end{tabular}%
}
\caption{\gld\ is comprised of RGB and depth point cloud images of 47 classes of objects in five high-level categories. It includes 8250 text and 4059 speech descriptions gathered with Amazon Mechanical Turk (AMT).}
\label{fig:dataset-example}
\end{figure}

The primary contributions of this paper are as follows:
\begin{enumerate*}
    \item We provide a freely available, multimodal, multi-labelled dataset of common household objects, with paired image and depth data and textual and spoken descriptions.
    \item We demonstrate this dataset's utility by analyzing the result of known grounded language acquisition approaches applied to transfer and domain adaptation tasks.
\end{enumerate*}

\begin{figure}[t]
    \centering
    \includegraphics[width=\columnwidth]{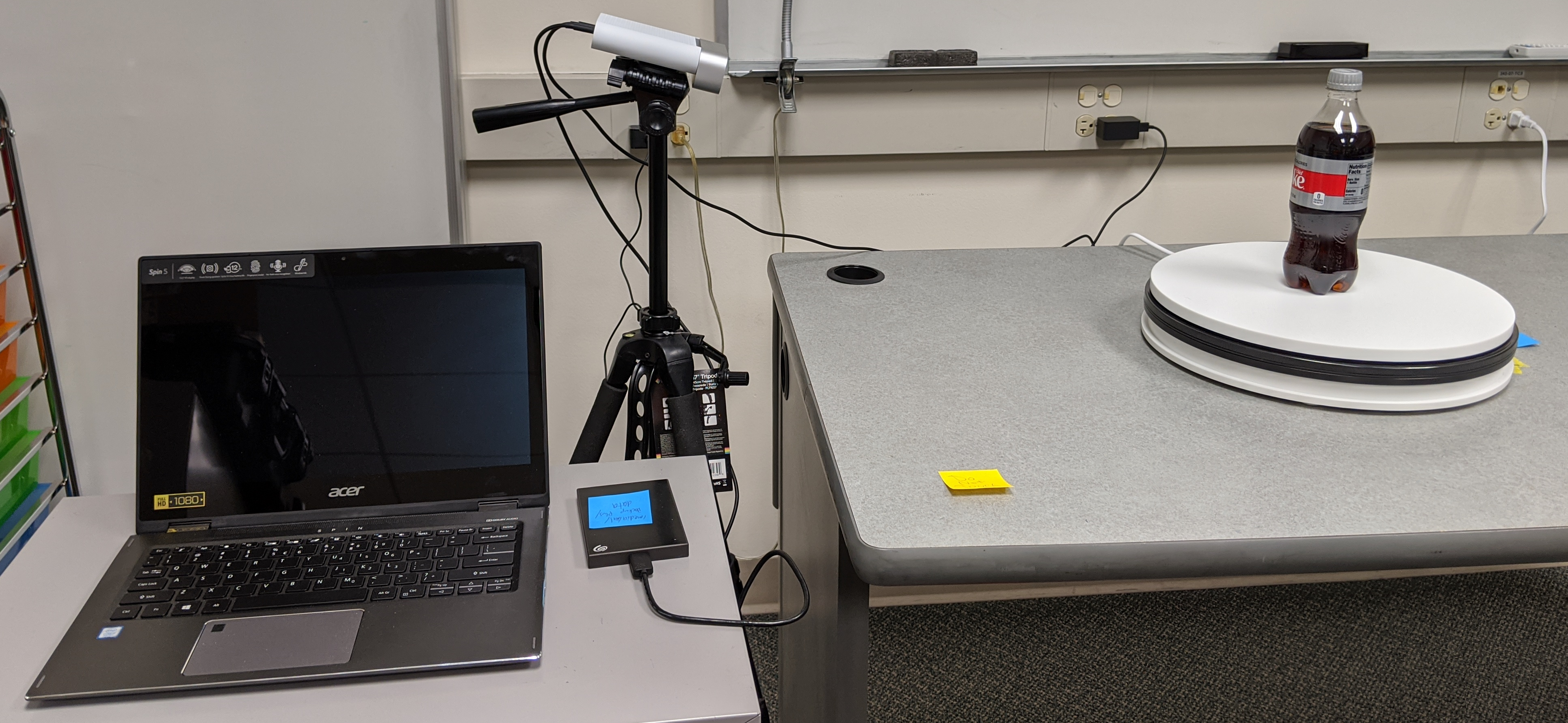}
    \caption{The data collection setup, inspired by~\citet{Lai2011uwrgbd}. An Azure Kinect (Kinect 3) is mounted on a tripod, pointed at the target object (in this case a soda bottle) on a white turntable. Image and depth data is collected as the object rotates on the turntable.
}
    \label{fig:data_collection}
\end{figure}

\section{The \gld\ Dataset}

\gld\ is a collection of visual and natural language data in five high-level groupings: food, home, medical, office, and tools. These were chosen to reflect and provide data for domains in which dynamic human-robot teaming is a near-term interest area. Perceptual data is collected as both images and depth while natural language is collected in both text and speech. There are 47 object classes spread across these high-level categories, each containing four to five instances of the object for a total of 207 object instances. During imaging, the objects are rotated on a turntable, allowing us to select four representative frames from different angles for a total of 825 views. For example, within the food category there is an apple class with five instances of apples, each with four distinct frames. Complete depth video of the objects from all angles is also available, but only these frames are labeled with multiple natural language descriptions in both text and speech. \gld\ is available on GitHub \textit{https://github.com/iral-lab/UMBC\_GLD}.


\begin{table}[h!]
\centering
\caption{Classes of objects in \gld.}
\label{table:categories}
\begin{tabular}{cp{5.333cm}}
\toprule
\textbf{Topic}   & \multicolumn{1}{c}{\textbf{Classes of Objects}} \\ \midrule
food    & {\it potato, soda bottle, water bottle, apple, banana, bell pepper, food can, food jar, lemon, lime, onion} \\
home    & {\it book, can opener, eye glasses, fork, shampoo, sponge, spoon, toothbrush, toothpaste, bowl, cap, cell phone, coffee mug, hand towel, tissue box, plate} \\
medical & {\it band aid, gauze, medicine bottle, pill cutter, prescription medicine bottle, syringe} \\
office  & {\it mouse, pencil, picture frame, scissors, stapler, marker, notebook}\\
tool    & {\it Allen wrench, hammer, measuring tape, pliers, screwdriver, lightbulb}\\ \bottomrule
\end{tabular}
\end{table}

\subsection{Accuracy of Speech Transcriptions}
\label{speech_accuracy}

Obtaining accurate transcriptions of speech in sometimes noisy environments is a significant obstacle to speech-based interfaces~\cite{li2013investigation}. In creating \gld\ we used the popular Google Speech to Text API,
chosen because it is widely available, easy to use, and not tied to a specific domain or hardware setup. However, the resulting transcriptions are therefore not tuned for optimal performance. For a particular use case, a more focused automatic speech recognition (ASR)  system could be used on the sound files included in the dataset. In order to understand the degree to which learning is affected by ASR errors, 100 randomly selected transcriptions were evaluated on a 4-point scale (see \cref{table:rating}). These descriptions were also manually transcribed (see \cref{table:example_ratings} for examples). Of those, 77\% are high quality, i.e., `perfect' or `pretty good,' while 13\% are rated `unusable.'


\begin{table}[t]
\caption{Human ratings of 100 automatic transcriptions. These ratings are designed strictly to assess the accuracy of the transcription, not the correctness of the spoken description with respect to the described object.}
\label{table:rating}
\begin{tabular}{clc} \toprule 
Rating & \multicolumn{1}{c}{Transcription Quality Guidelines} & \# \\ \midrule
1      & wrong or gibberish / unusable sound file      & 13 \\ 
2      & slightly wrong (missing keywords / concepts)  & 10 \\ 
3      & pretty good (main object correctly defined)   & 14 \\ 
4      & perfect (accurate transcription and no errors) & 63 \\ \bottomrule
\end{tabular}
\end{table}


In order to evaluate the replicability of the human-provided ratings in \cref{table:rating}, two subsets of these ratings was evaluated using Fleiss' kappa~\cite{fleiss1971measuring} ($\kappa$) to measure inter-annotator agreement across three raters. In both trials, $\kappa\approx 0.6$, representing moderate/substantial agreement among the raters. Although a higher agreement would be preferable, we observe that disagreement was never more than one unit between the raters, most commonly two ratings of 1 and one rating of 2.
As Fleiss' kappa does not incorporate concepts such as ``near agreement,'' for larger datasets, a weighted kappa statistic may be more appropriate. 

%
%

To get a more detailed understanding of transcription accuracy, we compare the ASR transcriptions and the human-provided transcriptions using the standard NLP metrics of Word Error Rate (WER) and Bilingual Evaluation Understudy (BLEU) score. Word Error Rate is recognized as the \textit{de facto} metric for automatic speech recognition systems, as WER strongly influences the performance of speech~\cite{cavazza2001empirical,park2008empirical,saon2006effect}. WER measures the minimum-edit distance between the system's results, the \textit{hypothesis}, and manually transcribed text, the \textit{reference}. WER is typically calculated as the ratio of word insertion, substitution, and deletion errors in a hypothesis to the total number of words in reference~\cite{mccowan2004use}. We evaluate WER for the same subset, and the mean WER per transcription is approximately 21.3\%. Out of 100 transcriptions, 42 had a zero error rate, meaning the reference and hypothesis match exactly.

BLEU is a closeness metric inspired by the WER metric. The main idea is to use a weighted average of variable length phrase matches against the references~\cite{papineni2002bleu}. BLEU scores are widely used to measure the accuracy of language translations based on string similarity; we adopt this system to evaluate the goodness of transcriptions. BLEU scores are calculated by finding $n$-gram overlaps between machine translation and reference translations. In our analysis, we use $N=4$.
\begin{equation}
    BLEU = \min\Bigg(1, \frac{output-length}{reference-length}\Bigg)\Bigg(\prod_{i=1}^{N}precision_i\Bigg)^\frac{1}{N}
    \label{eq:bleu_formula}
\end{equation}

\begin{figure}
\begin{center}
\includegraphics[width=0.99\columnwidth]{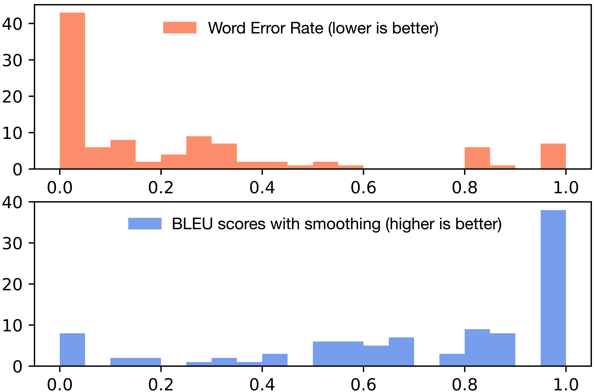}
\end{center}
\caption{Distribution of Word Error Rate and BLEU scores of automated compared to manual transcriptions. For higher-quality transcriptions (rated 3 or 4), the mean WER drops to 0.09, and the mean BLEU score rises to 0.83.}
\label{fig:bleu_wer_distribution}
\end{figure}

\noindent To mitigate the tendency of the BLEU score to penalize longer sentences, we apply a smoothing function while calculating scores and add 1 to both numerator and denominator while calculating precision~\cite{lin2004automatic}. We find that the mean BLEU score of the same subset of 100 transcriptions is 0.71, and Figure~\ref{fig:bleu_wer_distribution} shows the distribution.

\subsection{Comparative Analysis}

We analyze both the text and speech descriptions for the number of words used as well as mentions of color, shape, and object name. Since both modes are used to interface with robots, we wish to find any similarities or differences that might inform system design techniques for grounded language models. One of the learning targets in grounded language acquisition is to learn attributes of physical objects (as identified by natural language) such as color, shape, and object type~\cite{PillaiMatuszekAAAI2018}. These categories are limited because they are expert-specified and prescribed; the \gld\ dataset is intended to support learning of unconstrained, ``category-free''~\cite{richards2019learning} linguistic concepts. This would allow learning of attribute terms such as ``white'' or ``cylindrical,'' but also unexpected concepts such as materials (e.g., ``ceramic coffee mug'').

To that end, we analyze the natural language descriptions and find that color, shape, and object names often appear in natural language descriptions of images. We apply a list of 30 common color terms from large language corpora and compared each description to see if it included one of these terms~\cite{mylonas2015colorterms}. Similarly, we use a vocabulary list of shape terms to count how many descriptions included shape descriptions. It is worth noting that shape descriptions are less well defined than colors and that a better vocabulary of shape descriptions would be helpful towards this kind of analysis. Finally, we consider how often descriptions contain object labels, which would allow them to be linked to external models.





\begin{figure}[t] \centering
\includegraphics[width=0.9\columnwidth]{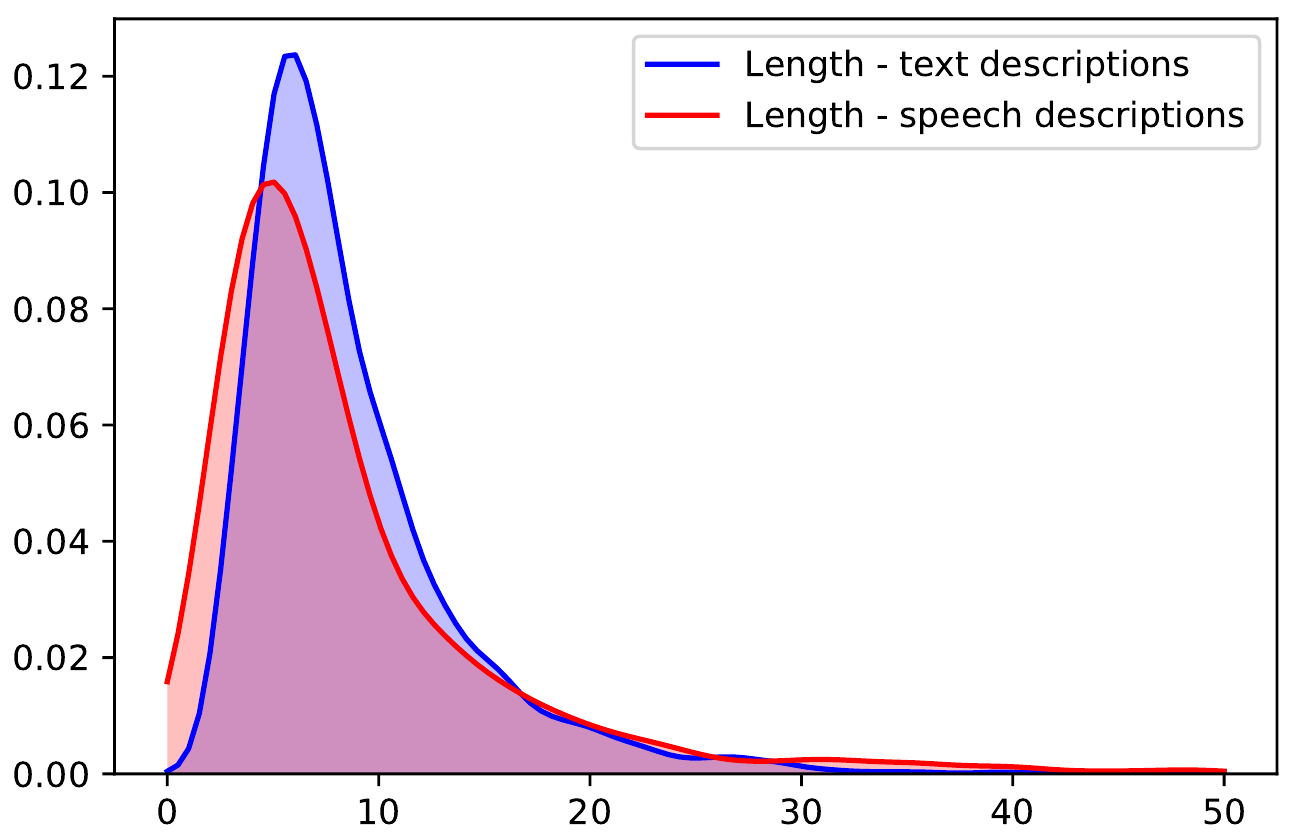}
\caption{Density Estimate plots of sentence length (the number of words) for the natural language descriptions with stop words removed. We found no significant difference in length of description between text and speech. $\mu_{textlength} = 8.38$, $\sigma_{textlength} = 5.14$; $\mu_{speechlength} = 8.72$, $\sigma_{speechlength} = 8.0$.}
\label{fig:kde_plots}
\end{figure}

Our initial hypothesis was that people would use more words when describing objects verbally than when typing, as it is lower effort to talk than to type.
When comparing description length, we balance the number of speech and text descriptions, using 4059 of each. However, we found no significant differences in the average length of descriptions between speech and text ($p\geq0.35$ using a Welch test or t-test) and in the distributions of mentions of color, shape, or object name between the two. While speech has slightly more average words per description, 8.72, compared to text at 8.38, when stop words are removed the averages are 4.52 and 4.38 respectively (see \cref{fig:kde_plots}).

The larger mean drop in the speech descriptions is likely due to the tendency of 
ASR systems to interpret noise or murmur utterances as filler words, the inclusion of which has been shown to detract from meaning~\cite{Engel2001,Engel2002,stolcke2017comparing}. Text descriptions are a more consistent length than speech, with a standard deviation of $\sigma_{textlength}=5.14$ words, versus $\sigma_{speechlength}=8.0$. When we remove stop words, the standard deviation is 2.58 for text and 3.96 for speech. 

As the \gld\ dataset contains 47 different classes, it is useful to note the class-wise differences in length of descriptions, in terms of the number of words. ~\Cref{table:mean_lengths_per_class} is a sample of some classes with interesting length differentials. In general, people tend to speak more, compared to what they type, when describing relatively complex objects, like ``syringe'', ``measuring tape'' 
or ``cell phone.'' On the other hand, speech descriptions for more basic objects such as ``banana'', ``spoon'', or ``eye glasses'' tend to be shorter even than their text descriptions. However, when taken all together the differential between text and speech length per object is not significant with an average of 0.14 more words in speech descriptions.

\begin{table}[b]
\centering \small
\caption{Mean length (number of words with stop words removed) in descriptions for selected categories, by description modality.}
\label{table:mean_lengths_per_class}
\begin{tabular}{ccc}
\toprule
\textbf{Category}  & \textbf{Length (Text)} & \textbf{Length (Speech)} \\
\midrule
measuring tape & 4.72   & 6.12 \\
syringe         & 3.8   & 4.75 \\
cell phone      & 4.17  & 4.98 \\
sponge          & 4.025 & 3.925 \\
food can        & 4.06  & 3.899 \\
scissors        & 4.98  & 4.79 \\
spoon           & 3.98  & 3.23 \\
eye glass       & 4.86  & 4.03 \\
banana          & 4.14  & 3.21 \\

\bottomrule
\end{tabular}
\end{table}

We use the Stanford Part-of-Speech (POS) Tagger\footnote{https://nlp.stanford.edu/software/tagger.shtml}~\cite{toutanova2003feature} to count the number of nouns, adjectives, and verbs in the descriptions. We are interested in evaluating these occurrences because they play a central role in defining groundings associated with any object. We find that the mean number of noun tokens per description is slightly higher in the text data (2.59) than the speech data (2.49). Similarly, the average number of adjectives per description is marginally higher for text data (1.25) compared to speech (1.17). Whereas the mean verb occurrence for text and speech are 0.52 and 0.62, respectively demonstrating the reverse trend.

\begin{table}[t]
\centering\small
\caption{Most frequent words in text (left) and speech (right).}
\label{table:frequencies}
\begin{tabular}{cc}
\toprule
\textbf{Token} & \textbf{\% Frequency} \\ \midrule
black          & 13.96               \\ 
object         & 12.66               \\ 
white          & 10.95               \\ 
blue           & 10.49               \\ 
red            & 11.76               \\ 
bottle         & 10.03               \\ 
yellow         & 9.61                \\ 
small          & 6.37                \\ 
used           & 6.19                \\ 
green          & 6.01                \\ 
pair           & 5.56                \\ 
plastic        & 4.53                \\ 
box            & 3.90                \\ 
silver         & 3.61                \\ 
metal          & 3.30                \\ 
pink           & 2.94                \\ 
picture        & 2.54                \\ 
orange         & 2.42                \\ 
large          & 2.53                \\ 
jar            & 2.08                \\ \bottomrule
\end{tabular}
\quad
\begin{tabular}{cc}
\toprule
\textbf{Token} & \textbf{\% Frequency} \\ \midrule
black          & 13.42                \\ 
white          & 12.31                \\ 
red            & 10.29                \\ 
blue           & 9.87                \\ 
bottle         & 8.86                \\ 
yellow         & 8.37                \\ 
object         & 6.87                \\ 
handle         & 6.62                \\ 
color          & 5.93                \\ 
green          & 5.56                \\ 
used           & 5.46                \\ 
small          & 4.67                \\ 
silver         & 4.06                \\ 
light          & 3.89                \\
box            & 3.81                \\ 
pair           & 3.74                \\ 
like           & 3.59                \\ 
plastic        & 3.57                \\ 
looks          & 3.07                \\ 
pink           & 2.58                \\  \bottomrule
\end{tabular}
\end{table}

Table~\ref{table:frequencies} shows the top twenty most frequent tokens in both categories. There is substantial overlap, as expected, since the same objects are being described. Words related to color are most commonly used to describe objects. People use more filler words when describing the objects using speech; for example, the word `like' appears 166 times in speech data whereas it was not significant in the text data. The word `used' appears frequently, typically used to describe the functionality of certain objects. Developing grounded language models around functionality for the analysis of affordances in objects is an important research avenue that our dataset enables, which is not possible in prior datasets that do not contain the requisite modalities.

\begin{figure*}[!ht]
    \centering
    \includegraphics[width=0.9\textwidth]{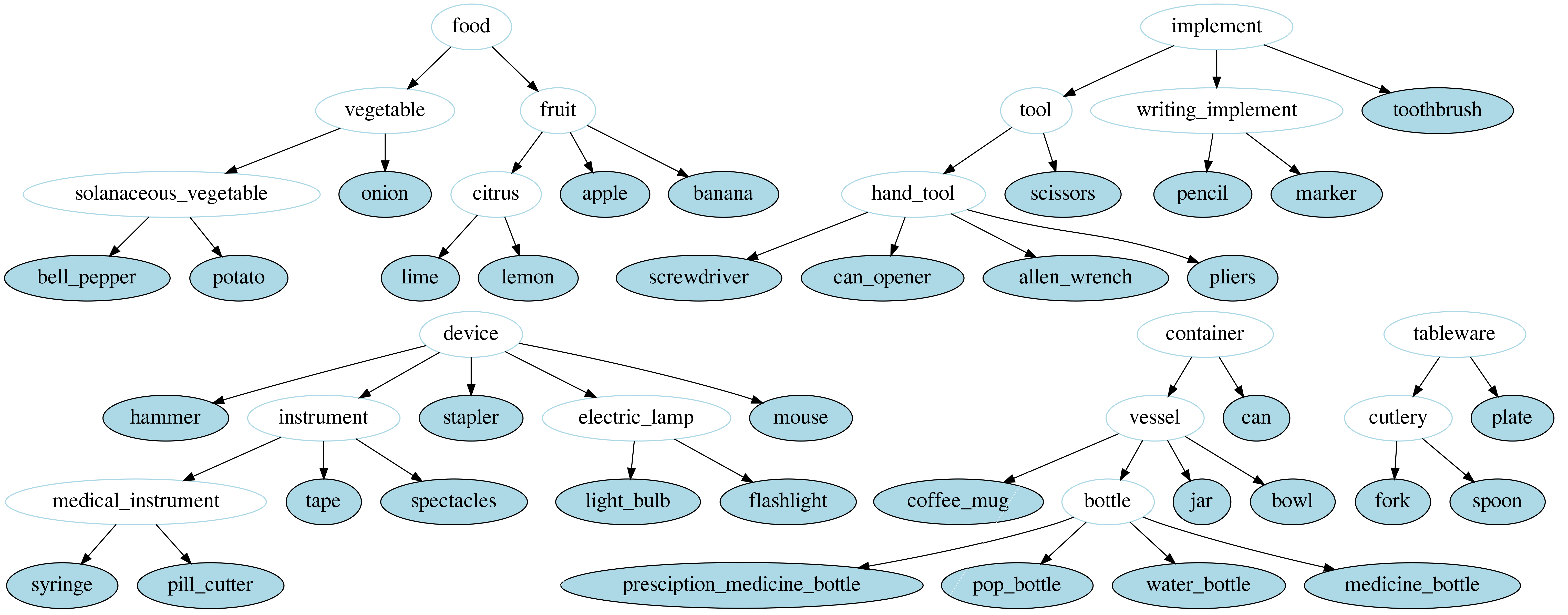}
    \caption{Subtrees highlighting object classes that appear in our dataset (light blue nodes), and the hierarchical structure of their related concepts as derived from WordNet~\cite{Miller95wordnet} (white nodes). These object class subtrees mirror similar category hierarchies reported in ImageNet~\cite{deng2009imagenet} and UW RGB-D~\cite{Lai2011uwrgbd} datasets.}
\end{figure*}

\subsection{Related Datasets}

Grounded language acquisition is in the unique position of requiring a dataset that combines sensory perception with language. These combined datasets are frequently handcrafted for the specific task that the research seeks to accomplish~\cite{Chen2008LearningTS,Roy2002LearningVG}, often leading to datasets with narrow applications. For example, CLEVR~\cite{Johnson2016clevr} was designed as a benchmark for question answering tasks. The dataset itself consists of scenes of colored toy blocks arranged in various positions. These scenes are annotated with the color, shape, size, and spatial relation to other objects within the scene. The simplicity of the scene along with the narrow scope of annotations in turn limits the type and complexity of questions that can be asked. As question answering and grounded language systems become more advanced, there is a need for our datasets to reflect real world scenes both in their composition and annotation. \gld\ achieves this by including real world objects of varying types and natural language.

A barrier to creating a dataset that includes speech is the high cost of collecting audio or transcribing it into a form that is usable by the intended system. \citet{Roy2002LearningVG} presents a grounded language system that can generate descriptions for targets within a scene of colored rectangles. The visual data for this task is easily generated, but for the speech descriptions, an undergraduate recorded 518 utterances over three hours. The audio from this collection was then manually transcribed into text. The manual audio transcription task can take anywhere between four and ten hours per hour of audio depending on the quality of the audio being transcribed and the final quality of the transcription~\cite{evers2015kwalitatief,evers2011past,weizs2019howlong}. We overcome this challenge by utilizing Speech-to-Text technology and evaluating the transcriptions for their quality as described in \cref{speech_accuracy}.

\noindent While not a grounded language dataset itself, it should be noted that the image collection of this work is heavily influenced by the University of Washington RGB-D dataset~\cite{Lai2011uwrgbd}. Both datasets contain large numbers of everyday objects from multiple angles. Our dataset is collected with a now-state of the art sensor which enables us to capture smaller objects at a finer level of detail (such as an Allen key, which is nearly flat against the surface). Additionally, we select objects based on their potential utility for specific human-robot interaction scenarios, such as things a person might find in a medicine cabinet or first aid kit, allowing learning research relevant to eldercare and emergency situations~\cite{beckerle2017human}.

\section{Dataset Creation}

In this section, we discuss the steps involved in collecting images, depth images, speech, and typed descriptions for objects in the \gld\ dataset. This includes the tools used and the crowd-sourcing activities required.

\subsection{RGB-D Collection}

Visual perception data is collected using a Microsoft Azure Kinect, colloquially known as a Kinect 3, using Microsoft's Azure Kinect drivers for Robot Operating System (ROS)\cite{koubaa2017robot}.\footnote{\url{https://docs.microsoft.com/en-us/azure/kinect-dk/}} The Kinect 3 is an RGB-D camera consisting of both a Time-of-Flight (ToF) depth camera and a color camera which enables it to capture high-fidelity point cloud data. We collected raw image and point cloud data from 47 classes of objects across the five categories of objects. Approximately five instances of each class are imaged for a total of 207 instances. \Cref{table:categories} shows the high-level topics as well as the specific classes that are collected within each topic. The dataset contains 207 90-second depth videos, one per instance, showing the object performing one complete rotation on a turntable. It also contains \textbf{825 pairs of image and depth point cloud from 207 objects,} consisting of manually selected representative frames showing different angles of each object (an average of 3.98 frames per instance).

The Kinect Azure depth camera uses the Amplitude Modulated Continuous Wave (AMCW) ToF principle. Near infrared (NIR) emitters on the camera illuminate the scene with modulated NIR light and the camera calculates the time of flight for the light to return to the camera. From this a depth image can be built converting the time of flight to distance and then encoded into a monochromatic depth image. ROS allows for the registration of the color and depth images, matching pixels in the color to pixels in the depth image, to build a colored point cloud of the scene. Point Cloud Library (PCL) \cite{Rusu_ICRA2011_PCL} passthrough filters are used to crop the raw point cloud to only include the object being collected and the turntable.



\subsection{Text Description Collection}

We collect our text descriptions using the popular crowdsourcing platform Amazon Mechanical TURK, or AMT. As described above, for each object instance, a subset of representative frames was manually chosen. For each task on AMT, these frames were shown for five randomly-chosen objects, each paired with a textbox. The AMT worker is asked to describe the object on the turntable in one or two short, complete sentences; they are specifically asked to not mention the turntable, table, other extraneous objects in the background. Each task was performed ten times, for a total of 40 text descriptions per instance. Removing objects for which there were representative frame errors, this allowed for the collection of \textbf{8250 total text descriptions}.

The purpose of taking images of objects from a variety of angles is to diversify what workers see. It is a known problem in vision systems that pictures tend to be taken from `typical' angles that most completely show the object; for example, it is rare for a picture of a banana to be taken end-on. This aligns with our motivation of creating a dataset of household objects to support research on grounded language learning in a human-centric environment: a robot talking to a person may have a partial view or understanding of an object, or vice versa. Thus we consider it essential to capture multiple views of objects in our dataset, and have those perhaps atypical views reflected in natural language descriptions.

\begin{figure*}[!ht]
    \centering
    \begin{subfigure}[b]{0.22\textwidth}
        \centering
        \includegraphics[width=1.0\columnwidth]{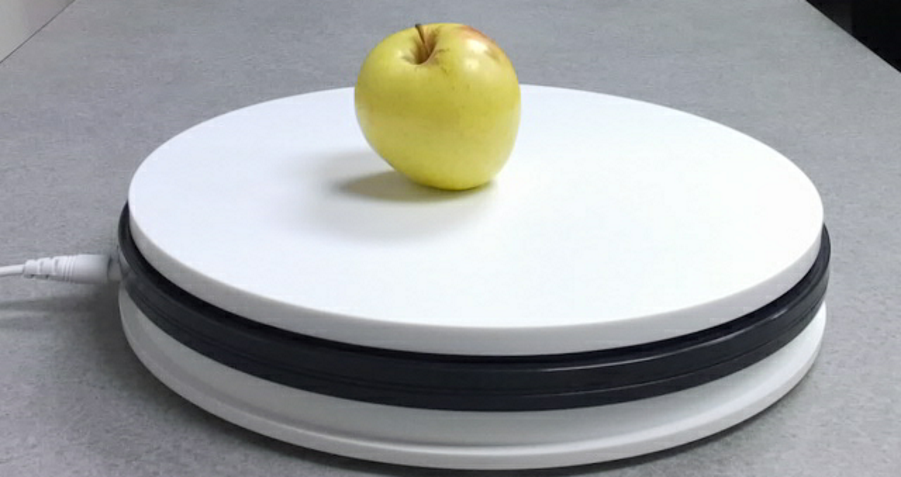}
        \caption{Apple image frame.}
    \end{subfigure}
    \begin{subfigure}[b]{0.24\textwidth}
        \centering
        \includegraphics[width=1.0\columnwidth]{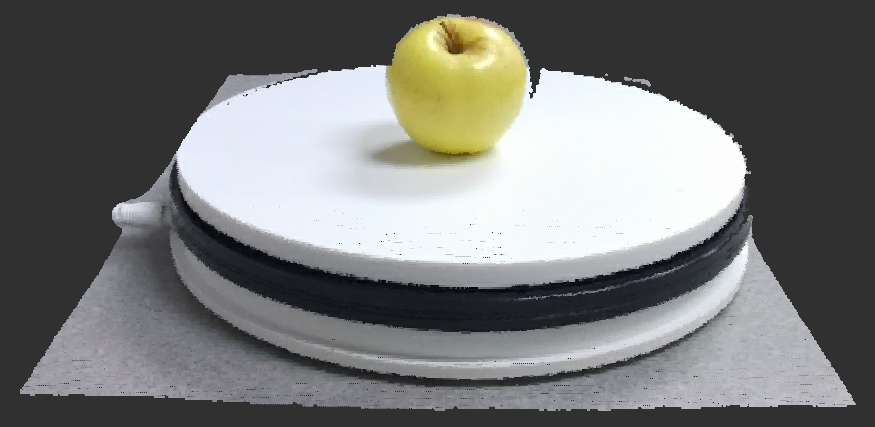}
        \caption{Apple depth point cloud.}
    \end{subfigure}
    \begin{subfigure}[b]{0.255\textwidth}
        \centering
        \includegraphics[width=1.0\columnwidth]{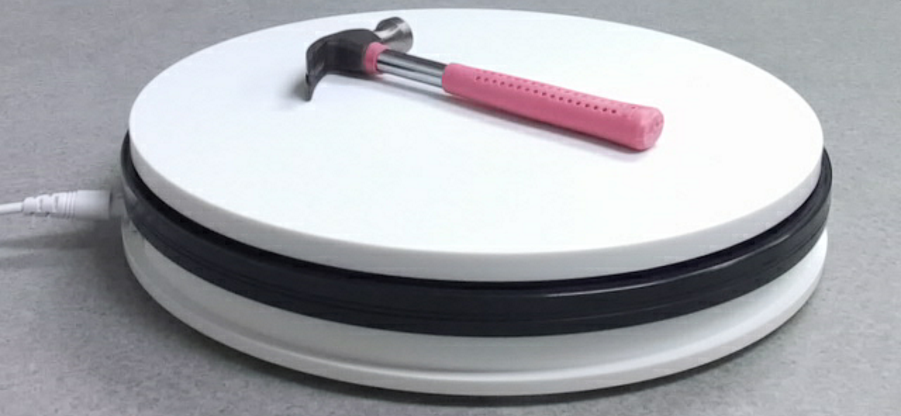}
        \caption{Hammer image frame.}
    \end{subfigure}
    \begin{subfigure}[b]{0.27\textwidth}
        \centering
        \includegraphics[width=1.0\columnwidth]{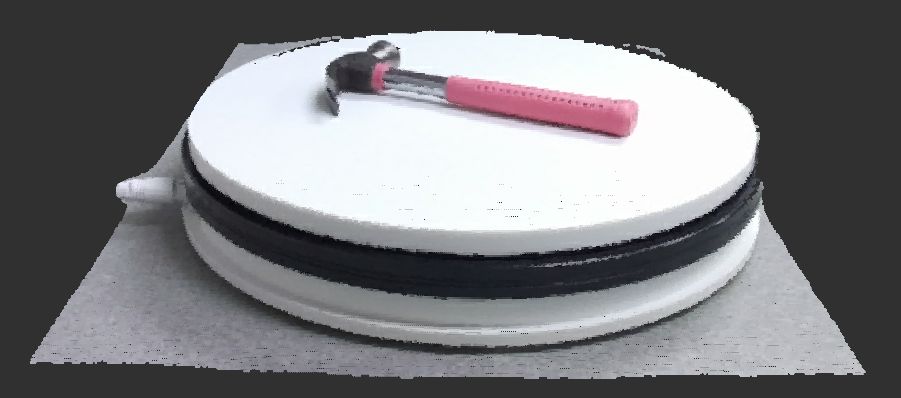}
        \caption{Hammer depth point cloud.}
    \end{subfigure}
    \caption{Samples showing the alignment of the visual data in \gld. Each instance contains a stream of RGB image frames (taken as the object rotates on the turntable), as well as an aligned 3D point cloud capturing depth information. Note: the RGB images have been cropped from the full size for display here.}
\end{figure*}
\subsection{Speech Description Collection}

As speech interaction is becoming more common with current technologies, our dataset will allow researchers to design and test grounded learning solutions using this popular input modality. We collect audio data to capture the nuances between spoken and written natural language. It is common for people to restructure sentences before writing them, but while speaking, we do not have the liberty to re-frame or restructure them. Therefore, spoken sentences tend to be unplanned, less well framed or grammatically incorrect~\cite{redeker1984differences}. Humans support speech with body gestures, eye gaze, expressions or pitch of the voice, details that are missing in writing~\cite{jaimes2007multimodal}. Experienced writers may be able to overcome these differences while communicating but these people usually hold formal education~\cite{Miller2006Spoken}. 

To collect spoken natural language data we develop a user interface utilizing the MediaStream recording API.\footnote{\url{https://developer.mozilla.org/en-US/docs/Web/API/MediaStream_Recording_API}} The audio clips are stored in an Amazon S3 bucket\footnote{\url{https://aws.amazon.com/s3/}}, which is a cloud storage service. Workers can play the recorded audio and if not satisfied can record it again. A similar approach is reported in recent work~\cite{lane-2010-naacl,lee-2015-isca} to collect data using web-based and mobile application-based systems. We embed the interface into AMT and the recorded audio files are collected from these tasks.

The task on Amazon Mechanical Turk had a simple interface, showing a single image with ``Record'', ``Play'', and ``Submit'' buttons. Each `task' had five such images, shown sequentially. In order to make the audio files compatible with ASR systems, missing metadata was added.
The audio files were converted to text using Google's Speech to Text API\footnote{\url{https://cloud.google.com/speech-to-text}}. A subset of these transcriptions was evaluated for quality, as explained in \cref{speech_accuracy} (some examples are shown in \cref{table:example_ratings}). This process resulted in a spoken-language dataset of \textbf{4059 verbal descriptions of 207 objects.}


\begin{table*}[t]
\centering\small
\caption{Some examples of transcription-quality ratings.
The transcriptions with the exact match are rated as 4.}
\label{table:example_ratings}
\begin{tabular}{ccll}
\toprule
\textbf{Rating} & \textbf{Class of Object Described} & \textbf{Google Speech to Text Transcription}                                                                                       & \textbf{Manual Transcription}                                                                                     \\ \midrule
1                & toothpaste                                                           & Institute best                                                                                                      & It's a toothpaste                                                                                                 \\ 
1                & spoon                                                                & did Persephone used to serving before                                                                               & \begin{tabular}[c]{@{}l@{}}this is spoon made up with \\ wood used for serving food\end{tabular}                  \\ 
1                & soda\_bottle                                                          & lovesick 100 African Buffalo                                                                                        & \begin{tabular}[c]{@{}l@{}}it is a plastic one and half \\ liter bottle of coke\end{tabular}                      \\ 
2                & stapler                                                              & \begin{tabular}[c]{@{}l@{}}this is the stuff inside mechanical \\ device  which joins Legends of paper\end{tabular} & \begin{tabular}[c]{@{}l@{}}this is a stapler it is a mechanical \\ device which joins pages of paper\end{tabular} \\ 
2                & can\_opener                                                           & \begin{tabular}[c]{@{}l@{}}emmanuel 10 opener with a \\ blue handle\end{tabular}                                    & \begin{tabular}[c]{@{}l@{}}A manual tin opener with a \\ blue handle\end{tabular}                                 \\ 
2                & hand\_towel                                                           & its a folded great owl                                                                                              & it's a folded gray towel                                                                                          \\ 
3                & shampoo\_bottle                                                       & what is a bottle of shampoo                                                                                         & that is a bottle of shampoo                                                                                       \\ 
3                & mouse                                                                & \begin{tabular}[c]{@{}l@{}}Addison black color Mouse can \\ be used in laptop or system\end{tabular}                & \begin{tabular}[c]{@{}l@{}}it is a black color Mouse can \\ be used in laptop or system\end{tabular}              \\ 
3                & coffee\_mug                                                           & Arizona white coffee mug                                                                                            & There is a white coffee mug                                                                                       \\ \bottomrule
\end{tabular}
\end{table*}

\section{Applications}

Grounded language is useful for many robotic tasks. Grounded language acquisition~\cite{alomari2017natural} is the general task of learning the structure and meanings of words based on natural language and perceptual inputs, usually visual but sometimes including other modalities such as haptic feedback or sound~\cite{thomason2018guiding}. Other tasks include navigation~\cite{unal2019visually} where a robot needs to either understand directions to get to a destination, or generate directions for someone or something else. Teaching and understanding tasks~\cite{chai2017teaching} as well as asking for help on tasks~\cite{tellex2014asking} are important areas of language that make interactions with robotic systems capable and productive. We focus on \textit{manifold alignment for grounded language acquisition} as our example use of \gld, because it is a grounded language acquisition task that highlights the multimodal nature of the data and the challenges unique to that setting.
Since grounded language acquisition is the most general and used to some degree in all of these applications, we choose to focus on an example that highlights the use of \gld\ towards a grounded language acquisition task.

\subsection{Example: Manifold Alignment}

We conduct a learning experiment to show how \gld\ might be used as a means to learn grounded language. We use manifold alignment~\cite{Richards2019NERC,wang2011heterogeneous,wang2008mainfoldalignment,andrew2013deepcca} with triplet loss~\cite{balntas2019triplets,schroff2015facenet} to embed the perceptual and language data from \gld\ into a shared lower dimensional space. Within this space, a distance metric is applied to embedded feature vectors in order to tell how well a particular utterance describes an image. Novel pairs can then be embedded to determine correspondence. Alternatively, inputs from either domain can be embedded in the intermediate space to find associated instances from the other domain.

For example, a picture of a lemon and the description ``The object is small and round. It is bright yellow and edible.'' should be closer together in the embedded space than the same picture of a lemon and the incorrect description ``This tool is used to drive nails into wood,'' since the latter description was used to describe a hammer. Through this technique, even novel vision or language inputs should be aligned, meaning that a new description of a lemon should still be closely aligned in the embedded space. We would additionally expect other similar objects, such as an orange, to be described in a somewhat similar way, allowing for potential future learning of categorical information.


\paragraph{Vision.}
The vision feature vectors are created following the work of Eitel et al~\cite{Eitel2015MultimodalDL}. The color and depth images are each passed through a convolutional neural network that has been pretrained on ImageNet~\cite{deng2009imagenet,russakovsky2015imagenet} with the last layer (which is used for predictions) removed so that the final layer is a learned feature vector. The two vectors, one from color and one from depth, are then concatenated into a 4096-dimensional visual feature vector.

\paragraph{Language.}
The language features are extracted using BERT~\cite{Devlin2019BERTPO}. Each natural language description is fed to a ``BERT-base-uncased'' pretrained model which gives us the individual embeddings of all the tokens in the description. We obtain the description embedding by performing average pooling over the word embeddings. Due to the contextual nature of its embeddings, BERT can differentiate between different meanings of the same word in different contexts. This results in semantically richer language features and a more meaningful embedding space. The resulting 3072-dimensional vector is taken as the description's language feature vector and associated to the visual feature vector of the frame it describes. Since the dataset contains ten natural language descriptions for each frame of an object, each visual feature vector is paired with ten different language feature vectors. The same process is repeated for the speech transcriptions.

\paragraph{Triplet Loss.}
The basic triplet loss function~\cite{balntas2019triplets,schroff2015facenet} uses one training example as an ``anchor'' and two more points, one of which is in the same class as the anchor (the positive), and one which is not (the negative). For example, while classifying images of dogs and cats the anchor might be a cat image, the positive would be a different cat image, and the negative would be an image of a dog. The loss function then encourages the network to align the anchor and positive in the embedded space while repelling the anchor and the negative. Typically the positive and negative instances are from the same domain as one another. However, they may also be from the same domain as the anchor in order for the the network to be internally consistent, or from the other domain to align the two networks. Each of the four cases is chosen uniformly at random during training for each training instance.

\paragraph{Negative Sampling.}
In our case there is no obvious conceptualization of positive and negative examples of language. Because language is not exhaustive, the fact that a description omits particular concepts does not mean that the omitted language would not describe the object (for example, describing a lemon as a ``yellow lemon'' does not make it a good counterexample for the concept `round'). Similarly, a description can include language that is accurate for a description of a different object or even a different class; even in our dataset, which focuses on deep coverage of a small number of classes, ``a round yellow thing'' can be a lemon, a light bulb, or an onion. This underpins the widespread difficulty of finding true negative examples for natural language processing.

One approach to solving this problem for natural language processing relies on a different use of feature embedding~\cite{PillaiMatuszekAAAI2018}. We calculate the cosine similarity between a language feature vector and all other language feature vectors within the training set. Vectors that are semantically similar will have a distance close to 1 while those further apart will be closer to 0. Therefore, we take the feature vector with the smallest cosine similarity as the negative and the largest as the positive. To get positive and negatives of images, we find the positive and negative of the image's associated language and then take the associated images of those instances. For anchors ($A$), positive instances ($P$), and negative instances ($N$), we compute embeddings of these points, then compute triplet loss in the standard fashion with a margin $\alpha=0.4$~\cite{schroff2015facenet} with Equation~\ref{eq:triplet_loss}:
\begin{equation} \label{eq:triplet_loss}
   \mathcal{L}(A, P, N) = \max(\lVert f(A) - f(P) \rVert_2^2 - \lVert f(A) - f(N) \rVert_2^2 + \alpha, 0),
\end{equation}
where $f$ is the relevant model for the domain of the input points.

\paragraph{Training.}
Two models are trained from the data, a text-based language model and a transcribed speech-based language model. The text model, T, is trained for 50 epochs on 6600 paired visual/text feature vectors and evaluated on a held out set of 1650 examples from \gld. A speech model, S, is trained from 3232 language vectors and their associated images and evaluated on a held out set of 828 transcriptions. A third model, T+S, is trained from both text and speech transcriptions to see how the combination of domains affects learning. We are interested in how training may be affected by differences in the way people describe objects through their word choice or structure. The automated transcription process also introduces noise into the speech descriptions, which has an effect on downstream performance.

\paragraph{Evaluation.}
We evaluate the network using the Mean Reciprocal Rank (MRR). The MRR is calculated by finding the distance of an embedding of a vector in one domain to all of the embeddings of the other domain, ordering them by Euclidean distance, and finding the rank of the testing instance in the ordered list. The reciprocals of these ranks are summed over the testing set and then averaged by the number of testing examples. When the number of testing examples is very high, the MRR can quickly approach zero even when the rank of the instance is in the top half of examples, rendering the metric difficult to interpret. To counteract this and to evaluate our model on a scenario that is more realistic to what it might be used for, instead of ranking the entire testing set, we rank a select few instances. The first ranking is only on the target, positive, and negative instances and the second is on the target and four other randomly selected instances. The first evaluation tests that the triplet loss is having an effect on the final model, while the second mimics what the model might do when faced with identifying objects in a cluttered scene.

In both cases, we test (1) identifying objects from language descriptions, and (2) choosing a description given an object. We use $x\rightarrow y$ to denote a test query from domain $x$, which identifies something to be returned from domain $y$. $\mathtt{L} \rightarrow \mathtt{V}$ therefore denotes the test case in which language is provided and an object must be chosen from its perceptual data, and vice versa.

The combined ``T + S'' model is evaluated three separate times. First, it is tested individually on held-out sets where V is drawn first from text, then from speech. It is then evaluated on the combination of the two held-out sets. Because we expect the speech and text to be similar, testing on the combination of them should perform better than testing on either evaluation set in isolation, and in fact, this is what we see.
Our results show that a grounded language model can be learned from the \gld\ data. In particular, when ranking the distances between an embedded target instance from one domain with a selection of embedded instances from the other domain, we expect the target to appear in the top half of the rankings, which we consistently see.

\paragraph{Discussion.}
We are investigating, first, whether using the \gld\ data to train these models in a manifold alignment experiment yields better performance than a random baseline, and second, how the performance of a language model trained on speech performs \textit{vs}. typed text.
Table~\ref{tab:ml_results} shows the results from our experiments. In all cases our experiments outperform the random baseline, where the target is expected to have a rank halfway down the ordering.
For the Triplet MRR and Subset MRR, we would thus like our models to perform better than $1/2$ and $1/3$ respectively since there are three objects in the triplet evaluation and five in the subset.
Table~\ref{tab:ml_results} shows, then, that our model has effectively achieved manifold alignment, aligning similar examples while repelling dissimilar ones.
The fact that the Subset MRR is above $1/3$, which is true in all cases (vision to language, language to vision, whether speech or text), tells us that our model is not just randomly selecting a target instance. So, when given a subset of instances, we can say that our model is able to select the target or rank the target highly.

The speech model performs marginally worse than the text model. This could be due to the smaller training dataset, but is probably fully explainable by the noise generated by the transcription process. Speech transcriptions allow for a one to one comparison of the two domains, but in future work, we will train a model over the raw audio. In a system that utilized one of these models, this would eliminate the need for a transcription step and may provide more accurate results since there may be tonal or inflection data that is lost in the transcription process.

When the combined model is evaluated against each individual domain, we find the performance drop that we expected. The minor drop in performance when evaluating on the combined held out set implies that the two domains are different enough that a model trained on both modalities together has more difficulty reaching the level of performance of a learned model trained on uniform data. While this is not unexpected, it is heartening that the performance drop is relatively small. Perhaps more importantly, the improvement on the combined test set demonstrates that the model is being trained to effectively interpret either spoken or typed input. 

Another interesting aspect of these results is the relative difference in the direction of the mapping, that is, which domain is chosen as the target. In particular, the $\mathtt{V} \rightarrow \mathtt{L}$ case outperforms $\mathtt{L} \rightarrow \mathtt{V}$. That is, selecting the associated language given a visual input is easier than selecting the associated vision given a language input. We suspect this observed behavior could be due to differences in the manifolds of the feature vector spaces. BERT is a highly pre-trained model~\cite{Devlin2019BERTPO}. The visual domain, however, is much more complex. Raw images, like an apple and a red ball, may be very close together while it is unlikely that their descriptions would be close together. BERT uses the context of a word to generate the embeddings so in this example even if the word ``red'' were used in both descriptions, the language embedding would be different. The raw images also all contain similar background scenery, the turntable, and the table.

\begin{table}[t]
    \centering\small
    \caption{Experimental results. Mean Rank Reciprocal for models trained on Text and Speech descriptions. For Triplet, MRR values above 0.5 demonstrate a successfully learned alignment between language and perceptual data; for Subset, MRR values above 0.33 demonstrate success. Triplet MRR is calculated from the target and a positively and negatively associated test data point, while Subset MRR is calculated from the target and a subset of four random test data points.}
    \begin{tabular}{cccc}
        \toprule
        \textbf{Model} & \textbf{Domain} & \textbf{Triplet MRR} & \textbf{Subset MRR} \\
        \midrule
         Text & $\mathtt{L} \rightarrow \mathtt{V}$    & 0.6658    & 0.4560 \\
        & $\mathtt{V} \rightarrow \mathtt{L}$         & 0.7342    & 0.4669 \\
        \midrule
        Speech & $\mathtt{L} \rightarrow \mathtt{V}$  & 0.6661    & 0.4391 \\
        & $\mathtt{V} \rightarrow \mathtt{L}$         & 0.7289    & 0.4562 \\
        \midrule
        T + S & $\mathtt{L} \rightarrow \mathtt{V}$    & 0.5954    & 0.4255 \\
              & $\mathtt{L} \rightarrow \mathtt{V}$ (Test on T) & 0.4670 & 0.4547 \\
              & $\mathtt{L} \rightarrow \mathtt{V}$ (Test on S) & 0.6651 & 0.4520 \\
        \midrule
              & $\mathtt{V} \rightarrow \mathtt{L}$         & 0.6762    & 0.4519 \\
              & $\mathtt{V} \rightarrow \mathtt{L}$ (Test on T) & 0.4389 & 0.4605 \\
              & $\mathtt{V} \rightarrow \mathtt{L}$ (Test on S) & 0.4587 & 0.4594 \\
       \bottomrule
    \end{tabular}
    \label{tab:ml_results}
\end{table}

\section{Discussion and Future Work}

%
%
%

In this paper we present \gld, a grounded language dataset of images in color and depth paired with natural language descriptions of everyday household objects in text and speech.
We aim to make this resource a useful starting point for downstream grounded language learning tasks such as spoken natural language interfaces for personal assistants and domestic service robots.

To demonstrate a potential use of \gld, we use the data to train models that perform heterogeneous manifold alignment.
We hope this dataset serves researchers as a resourceful starting point from which to explore many more techniques, model architectures, and algorithms that further our understanding of grounded language.
In particular, the inclusion of speech alongside written textual descriptions allows for side-by-side comparisons of the two domains grounded to physical objects, or for novel multimodal techniques involving all three domains of vision, text, and speech.

The idiosyncratic properties of \gld\ suggest many research questions for future study.
For example, we remark that \gld\ includes descriptions of the same object as observed from multiple angles.
One interesting question to explore, then, would be how to identify objects from a different perspective or when information is missing.
Another property of \gld\ is that some descriptions focus on the \textit{use} of the object (mostly agnostic to perspective), while others report perceptual qualities of the objects such as logos and other identifying features uniquely visible from the annotator's current perspective on that object.
This aspect could be incorporated into a human-robot interaction study that examines grounding language to objects in a physical as viewed by embodied agents from different vantage points. 

In the near term, we are interested in leveraging this dataset to train robots to understand natural language in order to perform tasks in a domestic context.
The inclusion of medical and kitchen supplies is critical to training a robot for tasks such as cooking, cleaning, and administering care.
As we work toward this goal, we anticipate creating an expanded catalog of items including the diverse ways in which people describe and talk about the wide variety of items they encounter every day.



\bibliographystyle{aaai}
\bibliography{aaai2020}
\end{document}